\documentclass[letterpaper, 10 pt, conference]{ieeeconf}  

\IEEEoverridecommandlockouts                              

\usepackage{graphicx} 
\usepackage{amsmath} 
\usepackage{bm}    
\usepackage{amssymb}  
\usepackage{caption}
\usepackage{subcaption}
\usepackage{float}
\usepackage[ruled,vlined]{algorithm2e}

\usepackage{color}
\usepackage[bookmarks=false]{hyperref}
\usepackage[noadjust]{cite}

\DeclareMathOperator*{\argmax}{argmax} 
\title{\LARGE \bf
Learning Insertion Primitives with Discrete-Continuous Hybrid \\Action Space for Robotic Assembly Tasks}

\author{Xiang Zhang, Shiyu Jin, Changhao Wang, Xinghao Zhu and Masayoshi Tomizuka$^{1}$
\thanks{$^{1}$ All authors are with the Department of Mechanical Engineering, UC Berkeley, CA 94720, USA.
{\tt\small \{xiang\_zhang\_98,jsy, changhaowang,zhuxh,tomizuka\}@berkeley.edu}}%
}

\begin{document}

\maketitle
\thispagestyle{empty}
\pagestyle{empty}

\begin{abstract}
This paper introduces a discrete-continuous action space to learn insertion primitives for robotic assembly tasks. Primitive is a sequence of elementary actions with certain exit conditions, such as ``pushing down the peg until contact". Since the primitive is an abstraction of robot control commands and encodes human prior knowledge, it reduces the exploration difficulty and yields better learning efficiency. In this paper, we learn robot assembly skills via primitives. Specifically, we formulate insertion primitives as parameterized actions: hybrid actions consisting of discrete primitive types and continuous primitive parameters. Compared with the previous work using a set of discretized parameters for each primitive, the agent in our method can freely choose primitive parameters from a continuous space, which is more flexible and efficient. To learn these insertion primitives, we propose Twin-Smoothed Multi-pass Deep Q-Network (TS-MP-DQN), an advanced version of MP-DQN with twin Q-network to reduce the Q-value over-estimation. Extensive experiments are conducted in the simulation and real world for validation. From experiment results, our approach achieves higher success rates than three baselines: MP-DQN with parameterized actions, primitives with discrete parameters, and continuous velocity control. Furthermore, learned primitives are robust to sim-to-real transfer and can generalize to challenging assembly tasks such as tight round peg-hole and complex shaped electric connectors with promising success rates. Experiment videos are available at 
\href{https://msc.berkeley.edu/research/insertion-primitives.html}{https://msc.berkeley.edu/research/insertion-primitives.html}.

\end{abstract}


\section{INTRODUCTION}

Nowadays, robotic assembly tasks have received significant attention from industry and academia. The assembly process involves complex contact dynamics and fiction, and it is often difficult to use traditional model-based control methods. Classically, human experts need to manually design a sequence of key points and motions for specific tasks, which is time-consuming and inefficient. 

Deep reinforcement learning (RL) methods have been introduced to robot assembly tasks and achieves promising results. Inoue \cite{inoue2017deep} first utilized RL to learn searching and insertion policies and the following works improved on several practical problems such as sample efficiency \cite{hoppe2020sample}, reward design \cite{schoettler2020deep}, sim-to-real gap \cite{son2020sim}, and robot stability during the training \cite{khader2020stability}. Recently, meta-learning \cite{schoettler2020meta} and transfer learning \cite{tanakatrans} approaches are employed to obtain more general assembly skills. Most RL-based methods directly control the low-level robot configurations such as position, velocity, external wrench. However, since robotic assembly tasks usually require a long horizon of control commands to finish, such action space design may increase the exploration difficulty and may not be efficient to learn. Therefore, the action space selection in learning robot assembly skills remains an open research question.

\begin{figure}
    \centering
    \includegraphics[width=240pt]{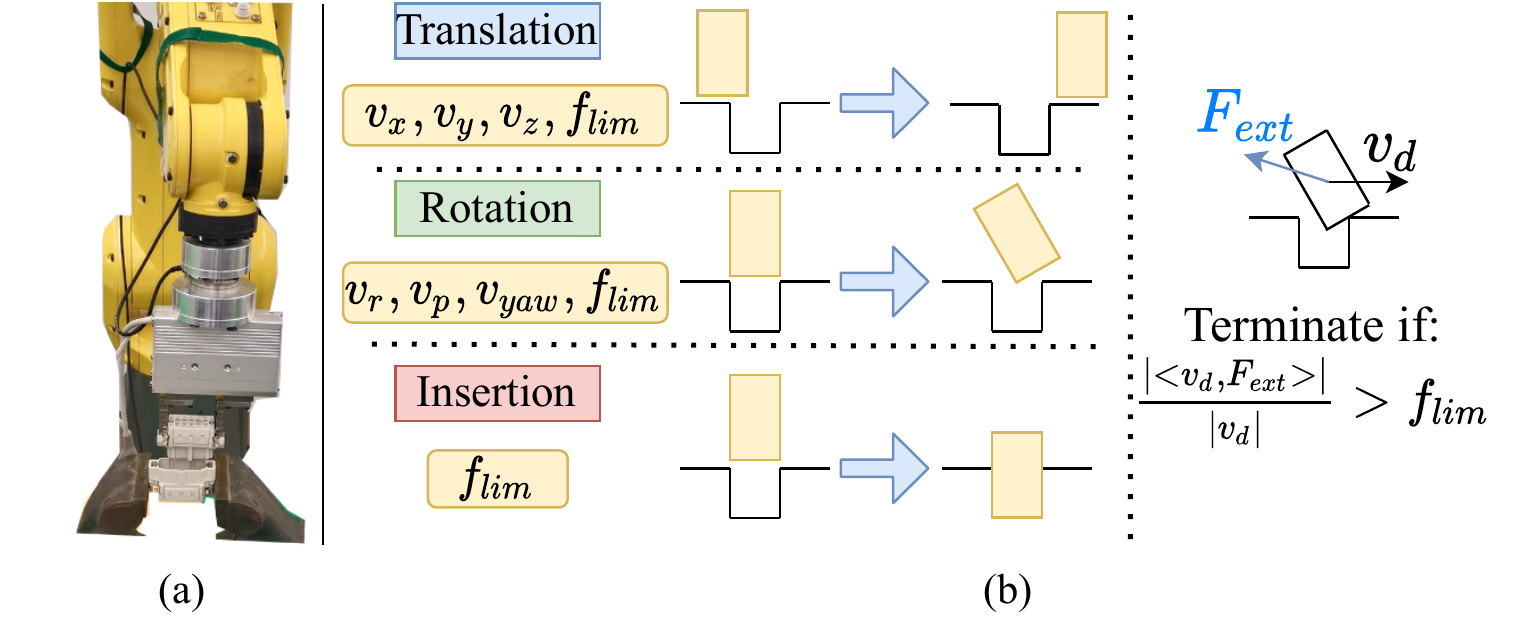}
    \caption{An overview of our proposed method: a) real-robot assembly task setup. b) three insertion primitive types: translation, rotation, insertion and their parameters. The contact force limit $f_{lim}$ serves as a terminal condition to stop the primitive when a contact happens.}
    \label{fig:overview}
\end{figure}

The recent development of primitive learning offers us a new insight into action space selection. Primitive is a sequence of elementary actions with certain exit conditions, such as pushing down the peg until contact and moving an object to a goal position. One manipulation task may consist of hundreds of control commands, while it can be represented by a few primitives. Thus, primitives can reduce the complexity of the task and significantly shorten the planning horizon. Furthermore, the design of primitives encodes human prior knowledge and provides additional semantics for better interpretability and robustness of the learned policy. The concept of primitive learning has been applied in a wide range of applications, including bimanual manipulation \cite{chitnis2020efficient}, video games \cite{delalleau2019discrete} and robotic soccer ball \cite{bester2019multi}. Vuong et al.\cite{vuong2020learning} introduced manipulation primitives to solve robotic assembly tasks and achieve good performance with real robots. However, their approach limits to discrete parameter selections for each primitive and is sensitive to hyperparameters such as the total number of parameter choices.
If the number is too small, discrete parameter selections cannot cover enough useful primitives to finish the task. If the number is too large, it makes the exploration difficult and reduces training efficiency. This issue limits their approach, and the experiments are conducted with relatively accurate hole pose ($\pm~1.5$ mm) 




In this paper, we formulate robot insertion primitives as parameterized actions \cite{masson2016reinforcement} that consist of discrete primitive types and continuous primitive parameters. As shown in Fig~\ref{fig:overview}(b), we consider three categories of insertion primitives: translation, rotation, and insertion. Their parameters are the Cartesian space velocity which indicates the translation and rotation direction, and a contact force threshold which stops the robot when contact happens. In addition, we further improve Multi-pass Deep Q-network (MP-DQN), which is the state-of-art RL method for parameterized actions, with twin Q-network and policy smooth to reduce the Q-network overestimation and proposed Twin-Smoothed MP-DQN (TS-MP-DQN) to learn robot insertion primitives.


Benefiting from learning primitives, our proposed approach is more efficient and has better generalization performance than using robot control commands as actions. Furthermore, we learn primitive parameters from a continuous space instead of selecting them from a discrete set, which improves the flexibility of learned primitives and yields better performance.
For validation, we compared our method with three baselines: MP-DQN with parameterized action space, primitives with discrete parameters, and continuous velocity control both in the simulation and on the real robot assembly tasks. From the experiment results, our proposed method can successfully learn insertion primitives in simulation and can be directly transferred to the real robot with a high success rate, even for unseen peg shapes. In addition, it outperforms all baselines in terms of the success rate and execution time.


The contributions of this paper are as follows:
\begin{itemize}
    \item We innovatively formulate robot insertion primitives as parameterized actions and use them as a hybrid action space to obtain robust insertion skills.
    \item We improved the MP-DQN method, the state-of-art RL method for parameterized actions with twin Q-network and policy smooth to reduce the overestimation of Q-network.
    \item We empirically evaluate the performance of our method by extensive experiments both in the simulation and the real world. Detailed comparisons with baseline methods are included.
\end{itemize}


\section{Reinforcement Learning and Parameterized Action Space}\label{RELATED WORK}
\subsection{RL for Discrete and Continuous Action Spaces}
The environment in RL are often modeled by the Markov Decision Process (MDP) $\mathcal{M} =(\mathcal{S}, \mathcal{A}, \mathcal{P}, r, \gamma)$, where $\mathcal{S}$ is the state space, $\mathcal{A}$ is the action space, $\mathcal{P}$ is the Markovian transition probability distribution, $r$ is the reward function, and $\gamma$ is the discount factor. The goal of RL is to find a optimal policy $\pi(a|s)$ which selects an action based on the current state to maximize the expected future return $\mathbb{E}\left[ \sum_{t} \gamma^t r_t \right]$.

Deep Q learning (DQN) \cite{mnih2013playing} is a model-free reinforcement learning algorithm. The intuition of Q learning methods is to estimate the future reward of taking action $a$ at state $s$, namely Q-value $Q(s,a)$:
\begin{equation}
    Q(s,a) = \mathop{\mathbb{E}}_{r,s'}[r+\gamma\max_{a'\in\mathcal{A}}Q(s',a')]
\end{equation}
where $r$ is the reward by taking action $a$ at state $s$, and $s'$ is the next state.
Once we have these Q-values, the greedy policy concerning  Q-values can be used to maximize the reward. The ideal Q-value $Q^*(s,a)$ is often approximated by a deep neural network $Q_\phi(s,a)$. Given a set of $\mathcal{D}$ transitions $(s,a,r,s',d)$ where $d$ indicates whether next state $s'$ is terminal, the Q learning target $y$ is calculated by $y(r,s',d) = r+\gamma(1-d)\max_{a'}Q_\phi(s',a')$ and the weights of the Q-network $\phi$ are updated using the following loss:
\begin{align}
    L_\phi =  \mathop{\mathbb{E}}_{(s,a,r,s',d) \sim D}[{(Q_\phi(s,a) - y})^2 ]
\end{align}

In the discrete action space, we have a finite number of action options, and a greedy policy can be easily obtained. However, for the continuous action space, the optimal action $a^*(s)=\max_{a}Q^*(s,a)$ cannot be efficiently computed. To overcome this problem, researchers proposed Deep Deterministic Policy Gradient (DDPG) \cite{lillicrap2015continuous} which adds a deterministic actor network $\mu_\theta(s)$ to directly output an action that approximately maximizes $Q_\phi(s,a)$. During the training, the actor network $\mu_\theta(s)$ is updated by performing gradient ascent to solve:
\begin{equation}
    L_\theta =  \mathop{\mathbb{E}}_{s \sim D}[- Q_\phi(s,\mu_\theta(s))] 
\end{equation}

\subsection{RL for the Parameterized Action Space}
Instead of modeling action space as fully continuous or discrete, the parameterized action space \cite{masson2016reinforcement} considers a hierarchical structure of actions, which contains high-level action $k$ and low-level action-parameters $x_k$. When interacting with the environment, the agent needs to first select a discrete action $k$ from a discrete set $[K] = \{1,2,\dots,K\}$ and then indicate its parameters $x_k \in\mathcal{X}_k$, where $\mathcal{X}_k$ is a continuous set for all $k \in [K]$. Formally, the parameterized action space $\mathcal{A}$ is defined as:

\begin{equation}
    \mathcal{A} = \{(k,x_k)|x_k\in \mathcal{X}_k\ for\ all\ k \in [K]\}
\end{equation}

One straightforward approach to learn with the hybrid action space is to relax into a fully continuous space. Specifically, the policy network outputs a continuous value for each discrete action. Then the discrete action can be selected by either picking the max value or sampling from a categorical distribution. One can find an example of this approach in \cite{hausknecht2015deep}. However, this approach does not utilize the special structure of the hybrid action space and might be more complex for learning. 

Another approach is to update the discrete action and continuous parameters separately. Xiong et al. proposed Parameterized Deep Q-Networks (P-DQN) \cite{xiong2018parametrized}, which modifies Q-network to include both the discrete action $k$ and its continuous parameters $x_k$:
\begin{equation}
    Q(s,k,x_k) = \mathop{\mathbb{E}}_{r,s'}[r+\gamma\max_{k'\in[K]}\sup_{x_{k'}\in\mathcal{X}_{k'}}Q(s',k',x_{k'})]
\end{equation}
and its actor network $\mu_k(s) $ approximates optimal parameters $\argmax_{x_k\in\mathcal{X}_k}Q(s,k,x_k)$ for discrete action $k$. Essentially, P-DQN utilizes the deterministic actor network from DDPG to output continuous actions and the Q-network from DQN to obtain the discrete action. Thus, it can be viewed as a combination of these two methods.

Since the input of Q-network consists of a joint action-parameter vector over all discrete actions, the Q-value of one discrete action is also related to the parameters of other discrete actions, which may not be reasonable and could lead to the false gradient in policy update. Multi-pass deep Q-networks (MP-DQN) \cite{bester2019multi} addressed this problem by zeroing out all unrelated action parameters when calculating Q-value with a multi-pass structure and achieved state-of-the-art performance on parameterized actions benchmark tasks.
\section{PROPOSED APPROACH}\label{PROPOSED APPROACH}

\subsection{Parameterized Insertion Primitives} \label{Insertion Primitives}

Similar to \cite{vuong2020learning}, we define one insertion primitive as the robot end-effector's desired motion, which consists of the desired movement command and a stopping condition as parameters. However, in their method, they proposed to use a set of discrete primitives parameters for each primitive and treat primitives as discrete actions. Thus, the agent only has limited choices of primitives from a finite set, which may result in sub-optimal and less flexible policies if the set is too small. In \cite{vuong2020learning}, the authors designed 91 primitives in total to include enough primitives. Nevertheless, it is a large discrete action space to search and could make learning more challenging.


In this paper, we formulate insertion primitives as parameterized actions and learn both the primitive type and its continuous parameters by RL to address these problems. As shown in Fig~\ref{fig:overview}(b), we consider three types of insertion primitives: \emph{translation}, \emph{rotation}, and \emph{insertion}, and their parameters are end-effector velocity commands and contact force limits. The details are introduced below.

\textbf{Translation primitive:} The translation primitives let the robot translate the peg in the Cartesian space without rotation. The parameters are $(v_x,v_y,v_z,f_{lim})$, where $(v_x,v_y,v_z)$ is the translation velocity representing the moving direction and speed, and $f_{lim}$ denotes the contact force limit. The velocity command sent to the robot is $v_{d}=(v_x,v_y,v_z,0,0,0)$, and the primitive should stop when the projection of contact force $F_{ext}$ in moving direction $v_d$ is greater than $f_{lim}$ : $\frac{|<v_d,F_{ext}>|}{|v_d|}>f_{lim}$. The moving distance threshold for this primitive is 25~mm, and the robot will stop and choose another primitive when reaching this threshold.

\textbf{Rotation primitive:} The rotation primitive rotates the peg and changes its orientation. It is parameterized by $(v_r,v_p,v_{yaw},f_{lim})$, where $(v_r,v_p,v_{yaw})$ is the rotation velocity in roll, pitch, yaw directions. The moving distance threshold is $4^\circ$ for rotation primitives. The maximum velocity $v_{max}$ is $ 0.5~cm/s~or ~rad/s$ and $f_{lim}$ is bounded by $[0,5]~N$.

\textbf{Insertion primitive:} For the insertion primitive, we expect the agent to use it when the peg is aligned to the hole by using the previous two primitives. Thus, the desired velocity command is $(0,0,-v_{max},0,0,0)$, which pushes the peg downwards with full speed. The only parameter for this primitive is $f_{lim}$ and the moving distance threshold is the same as the translation primitive. One thing to note is that there are some overlaps between the translation primitive and the insertion primitive. The reason for us to add an extra insertion primitive is to separate the searching phase and the insertion phase. Moreover, our preliminary experiments show adding insertion primitive can increase the success rate and reduce the execution time.

In the parameterized action formulation $a = (k,x_k)$, we have discrete primitive types $k = 0,1,$ and $2$ representing translation, rotation and insertion, respectively. The parameters $x_k$ for each discrete primitive type are shown in the previous discussion. We use these parameterized insertion primitives as the action space in RL.


\subsection{Twin Smoothed MP-DQN}
Though MP-DQN can achieve good performance in many scenarios. It is vulnerable for two common failure modes due to the DDPG-style when learning parameterized actions.
First, since DDPG uses a deterministic actor to approximate optimal continuous actions, it may explore errors of Q-network, which give the agent high values for false actions. Second, since only one Q-network is utilized during the training of DDPG, the learned Q-function may dramatically overestimate Q-values at some points of training. Inspired by TD3\cite{fujimoto2018addressing}, we advance the MP-DQN algorithm with clipped double Q-learning and target policy smoothing to address these two problems. We name this method as Twin-Smoothed Multi-pass Deep Q-network (TS-MP-DQN).

\begin{figure}
    \centering
    \includegraphics[width=200pt]{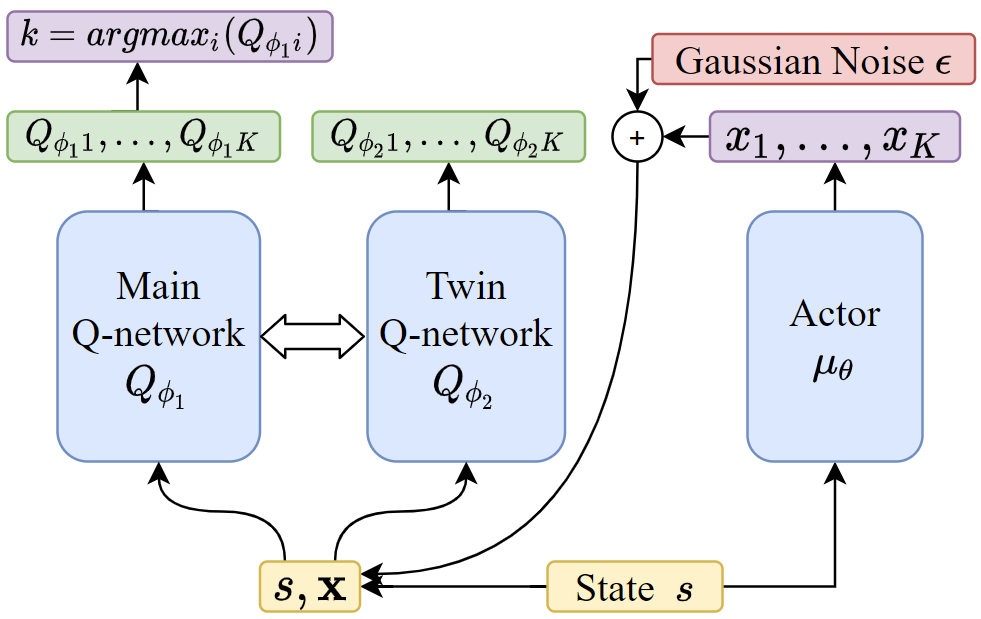}
    \caption{The policy network architecture}
    \label{fig:network}
\end{figure}

As depicted in Fig~\ref{fig:network}, the whole policy consists of three networks: a main Q-network $Q_{\phi_1}(s,k,x_k)$ to evaluate the Q-value by taking action $(k,x_k)$ at state $s$, a twin Q-network $Q_{\phi_2}(s,k,x_k)$ to reduce value overestimation, and an actor network $\mu_\theta(s)$ to approximate optimal primitive parameters $\mathbf{x} = (x_1,x_2,\dots,x_K)$ for all discrete actions. To infer the parameterized action $(k,x_k)$, we first calculate all primitive parameters $\mathbf{x}$ based on the state $s$. Then we use $(s,\mathbf{x})$ to obtain Q-values $(Q_{\phi_1 k} = Q_{\phi_1}(s,k,x_k),k\in[K])$ with the main Q-network and select the primitive type which has the highest Q-value. Like DQN-style methods, each network has one target network to provide a slow-moving target during the training. We denote their weights as $\phi_1',\phi_2',$ and $\theta'$.

To solve the first problem, we smooth the output of the actor network $\mu_\theta$ by adding a clipped Gaussian noise, when collecting transition pairs and calculating Q-learning targets:
\begin{equation}
    \mathbf{x}(s) = clip(\mu_\theta(s) + clip(\epsilon,-c,c), \mathbf{x}_{low}, \mathbf{x}_{high}), \epsilon \sim \mathcal{N}(0,\sigma)
\label{equ:smooth_policy}
\end{equation}
where $\sigma$ is the standard deviation of the Gaussian noise, $c$ bounds the Gaussian noise to a region, and  $\mathbf{x}_{low}, \mathbf{x}_{high}$ are the parameter limits. This trick can be viewed as a regulation to improve the stability of the algorithm. On the one hand, it is more difficult for the smoothed policy to utilize errors of the Q-network, which give incorrect value peaks for some specific action parameters. On the other hand, it encourages exploration and yields better results.

To address the second problem, another modification is to use the twin Q-network to reduce the Q-value overestimation. Following the clipped double Q-learning, we use the smaller Q-value given by two target Q-networks when calculating Q-learning targets:
\begin{align}
    y(r,s',d) = r+\gamma(1-d)\min_{\phi_1', \phi_2'}\max_{k'}Q_\phi'(s',k',x_{k'}(s'))
\end{align}
where $x_{k'}(s')$ is the approximated optimal action-parameter $\argmax_{x_{k'}}Q_\phi'(s',k',x_{k'})$ and is obtained from the smoothed target actor network output $\mathbf{x}'(s')$. Then both the main Q-network $Q_{\phi_1}$ and the twin Q-network $Q_{\phi_2}$ are updated by minimizing the MSE loss:
\begin{equation}
L_{\phi_i} =  \mathop{\mathbb{E}}_{(s,(k,x_k),r,s',d) \sim D}[{(Q_{\phi_i}(s,k,x_k) - y})^2 ], i\in[1,2]
    \label{eqn:MSE_loss}
\end{equation}

Finally, the actor network is updated by maximizing the main Q-network $Q_{\phi_1}$ output using the following loss:
\begin{equation}
    L_{\theta} = - \mathop{\mathbb{E}}_{s \sim D}\sum_k^K Q_{\phi_1} (s,k,{x_k}(s))
    \label{actor_loss}
\end{equation}
where ${x_k}(s)$ is a part of the smoothed actor network output $\mathbf{x} = (x_1,x_2,\dots,x_K)$ obtained by~(\ref{equ:smooth_policy}). The details of TS-MP-DQN are summarized in Algorithm~\ref{algo}.



\begin{algorithm}
\SetAlgoLined
Initialize Q-networks $Q_{\phi_1}$, $Q_{\phi_2}$ and actor network $\mu_\theta$ with random weights $\phi_1,\phi_2,\theta$

Initialize target networks $\phi_1' \leftarrow\phi_1,\phi_2' \leftarrow\phi_2,\theta'\leftarrow\theta$

Initialize replay buffer $\mathcal{B}$

\For{iteration t = 1 to T}{
Select parameters for all primitives using~(\ref{equ:smooth_policy})

Select a random discrete primitive type $k$ with probability $\epsilon$,

otherwise $k = \argmax_k Q_{\phi_1}(s, k, x_k)$

Execute hybrid action $(k,x_k)$, receive next state $s'$ and reward $r$

Store transition $[s, (k,x_k), r, s']$ into $\mathcal{B}$

Sample a minibatch of transitions from $\mathcal{B}$

Calculate the clipped double Q-learning target:
$y(r,s',d) = r+\gamma(1-d)\min_{\phi_1', \phi_2'}\max_{k'}Q_\phi'(s',k',x_{k'}(s'))
$

Update two Q-networks to minimize the MSE loss~(\ref{eqn:MSE_loss})

Update the actor network using~(\ref{actor_loss})

Update target networks:

$\phi_i' \leftarrow \tau \phi_i + (1-\tau)\phi_i'$; 
$\theta' \leftarrow \tau \theta + (1-\tau)\theta'$
 }
 \caption{Twin smoothed MP-DQN}
 \label{algo}
\end{algorithm}


\subsection{Transfer to Different Peg Shapes}



After training on one specific peg shape, we have two networks in hand: the main Q-network $Q_{\phi_1}(s,k,x_k)$ which evaluates the discrete primitive type and its continuous parameters, and the actor network $\mu_\theta(s)$ which is obtained by maximizing $Q_{\phi_1}(s,k,x_k)$. We design two strategies to transfer the learned primitives to different peg shapes: First, directly transfer both the Q-network and the actor without fine-tuning. Second, fix the Q-network to retrain the actor network for short episodes and then fine-tune the Q-network. Our intuition for the later method is that the Q-network encodes a higher-level policy of choosing insertion primitives which can be transferable among different peg shapes.
\section{EXPERIMENTS}\label{EXPERIMENTS}
\subsection{Training details}
\subsubsection{Peg-in-hole Environment}

\hfill

\begin{figure}[http]
    \centering
    \includegraphics[width=150pt]{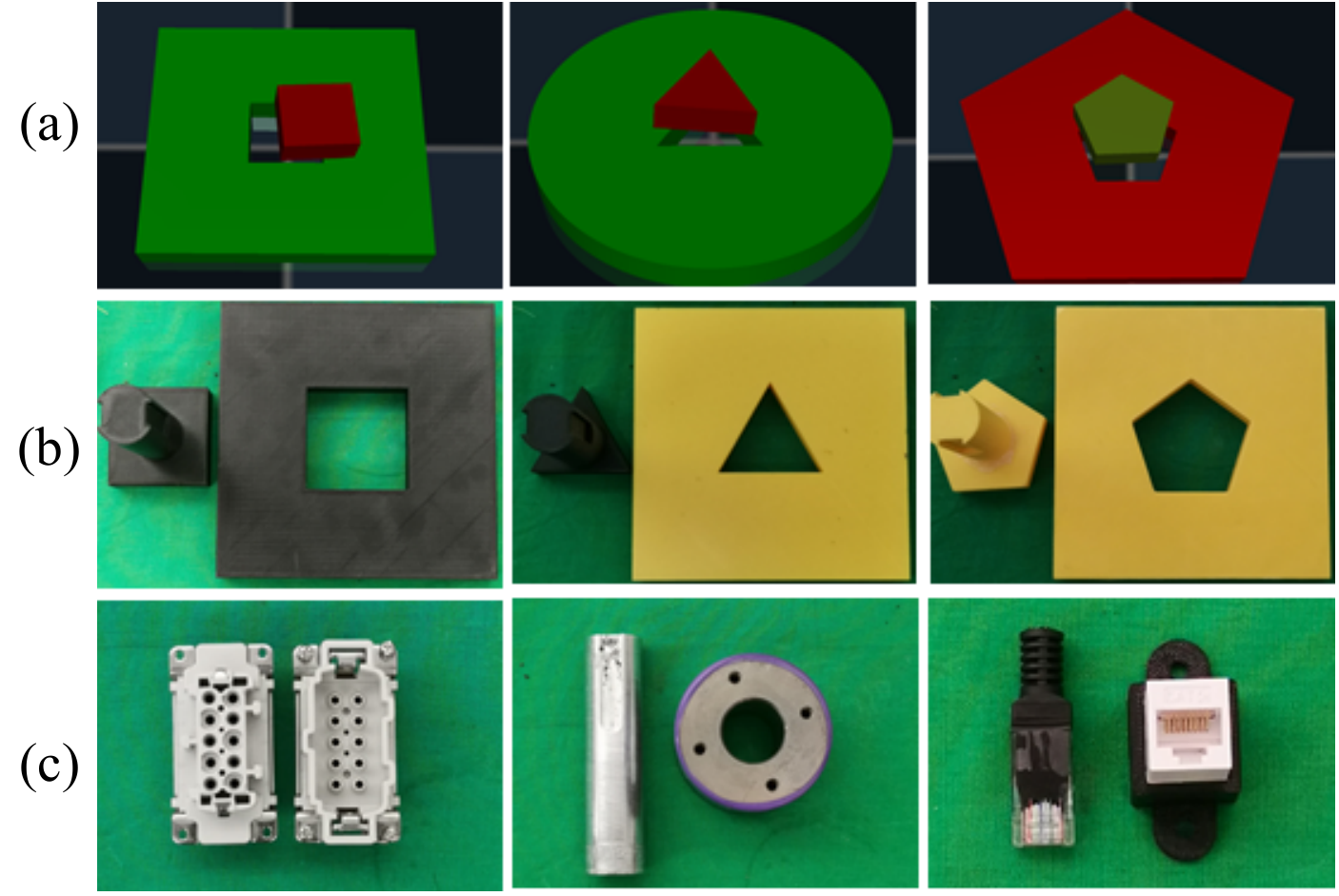}
    \caption{Real robot assembly tasks }
    \label{fig:assembly_tasks}
\end{figure}

\textbf{Simulation setup:} As shown in Fig~\ref{fig:assembly_tasks}(a), the simulation peg-in-hole environment consists of a peg and hole, and is build upon the Mujoco \cite{todorov2012mujoco} physics engine. There are three different peg-hole shapes in simulation, which are square, pentagon, and triangle peg-holes. We use the square-shaped peg-hole to learn insertion primitives and then test transfer learning performance on the other two shapes. Regarding the uncertainty in assembly tasks, a zero-mean Gaussian noise is added to the nominal hole pose $P_n$ to set the true hole pose $P_t$. We have uncertainty on $x,y$ directions of position and yaw direction of orientation and their standard deviation are $(2~mm,2~mm,0.5^\circ)$, respectively. The uncertainties are roughly bounded by $(\pm6~mm,\pm6~mm,\pm1.5^\circ)$ using the three-sigma rule. The peg reset position in $Z$ axis is $30$~mm above the hole and is uniformly sampled from $[-30,30]$~mm for $X,Y$ axes. The initial orientation in yaw direction is selected from $[-10^\circ,10^\circ]$ by a uniform distribution.

\textbf{State space:} The state space is defined as $s = (x,\Dot{x},F_{ext})$, where $x$ is the relative pose between the peg pose and the nominal hole pose, $\Dot{x}$ is the peg velocity, and $F_{ext}$ is the external wrench applied on the peg. 


\textbf{Reward function:} The reward function is defined as:
\begin{equation}
    r(s) = 10^{(3-{\Vert x - x_{goal}\Vert}_2)}
\end{equation}
Essentially, this reward function encourages the peg to get close to the goal location $x_{goal}$. When the peg reaches the goal, the agent will receive $1000$ reward, and the reward is close to 0 when the pose error is large.
\subsubsection{Baselines}

\hfill

We consider three baseline comparisons with our proposed method, which are pure MP-DQN using parameterized action space, a continuous action baseline using continuous velocity control command as action, and a discrete primitive baseline proposed in \cite{vuong2020learning}. The discrete primitives have six translation or rotation directions ($\pm X,\pm Y,\pm Z$), two velocity choices ($v=0.2$ or $0.5$), and four force thresholds ($f_{lim} = 1,2,3,+\infty$). Thus the total primitive numbers are 48 for translation and rotation, 4 for insertion. 

\subsubsection{Hyperparameters and Implementation}

\hfill

For the network architectures, our Q-network and actor network are two-layer Relu networks with 128 units. For the hyperparameters, we use a batch size of 128, the replay memory size of 10,000 and the discount factor $\gamma = 0.95$. The policy smooth parameter $\sigma = 0.2$ at the first episode and we linearly decrease it to $0$ after collecting 10,000 trajectories. The noise limit $c = e^{-2}$. 

Our algorithm is implemented based on the open source code from \cite{bester2019multi}. The continuous action baseline and the discrete primitive baseline are trained by SAC \cite{haarnoja2018soft} and DQN \cite{mnih2013playing} using implementations from RLkit \cite{rail-berkeley}.
\subsection{Simulation Experiments}
\subsubsection{Training on Square Peg-hole}

\hfill

\begin{figure}
    \centering
    \includegraphics[width=245pt]{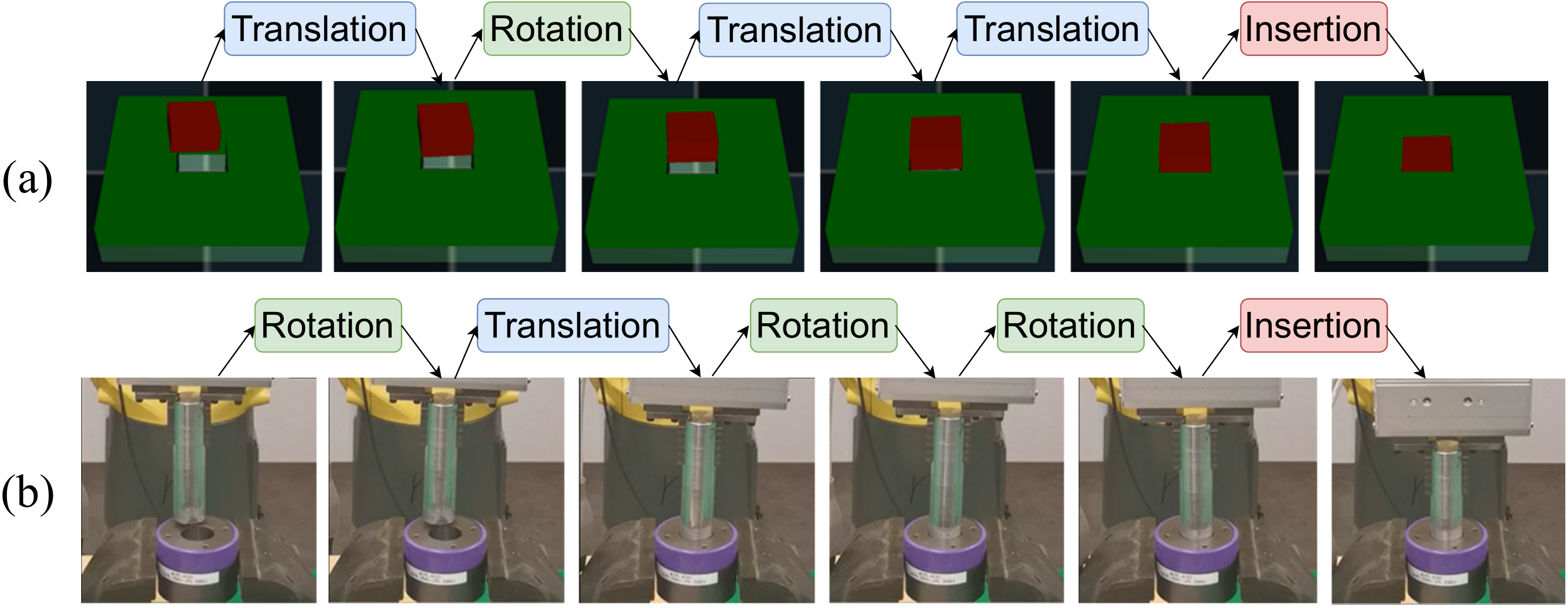}
    \caption{Snapshots of learned insertion primitives on a) square peg-hole in simulation b) tight round peg-hole on the real-robot}
    \label{fig:Snapshots}
\end{figure}
We first evaluate our method on a square peg-in-hole task in simulation. Fig~\ref{fig:Snapshots}(a) shows snapshots of the insertion process. We find that the insertion primitives learned by our method follow a particular ``tilt and translate" way to search for the hole location. The strategy employed by the agent is to first tilt the peg a little bit with a rotation primitive and then translate it downwards in the tilt direction using a translation primitive. The ``tilt and translate" strategy helps to establish contact between peg and hole, and find the actual hole pose. After that, the agent translates the peg in another direction to find both edges of the hole and then tilt back to align. Finally, the agent selects an insertion primitive to accomplish the assembly task.
\begin{table*}[http]
\begin{tabular}{c|ccc|ccc|c|cc}
         & \multicolumn{3}{c|}{H=10} & \multicolumn{3}{c|}{H=20} & \multicolumn{1}{c|}{} & \multicolumn{2}{c}{Transfer learning} \\
         \cline{2-10}
 & TS-MP-DQN & MP-DQN & \multicolumn{1}{c|}{Discrete} & TS-MP-DQN & MP-DQN & \multicolumn{1}{c|}{Discrete} & \multicolumn{1}{c|}{Continuous} & TS-MP-DQN &\multicolumn{1}{c}{MP-DQN}\\
 \cline{1-10}
Square   &$87.4\%$&$79.6\%$&$76.3\%$&$\bm{94.6\%}$&$92.0\%$&$91.0\%$&$87.0\%$\\
Triangle &$70.8\%$&$60.8\%$&$64.3\%$&$\bm{88.2\%}$&$82.4\%$&$83.6\%$&$66.7\%$&$\bm{92.8\%}$&$88.0\%$\\
Pentagon &$65.2\%$&$59.6\%$&$44.0\%$&$\bm{87.8\%}$&$82.4\%$&$82.3\%$&$50.3\%$&$\bm{83.4\%}$&$\bm{83.2\%}$\\                  
\end{tabular}
\caption{Mean success rate in the simulation over 5 random seeds. Each method is evaluated for 100 trials on one task. $H$ is the maximum primitive number the agent can use or the maximum horizon. Discrete is short for discrete primitive baseline, and Continuous denotes continuous control baseline.}
\label{tab:sim_success_rate}
\end{table*}

\begin{figure}
    \centering
    \includegraphics[width=240pt]{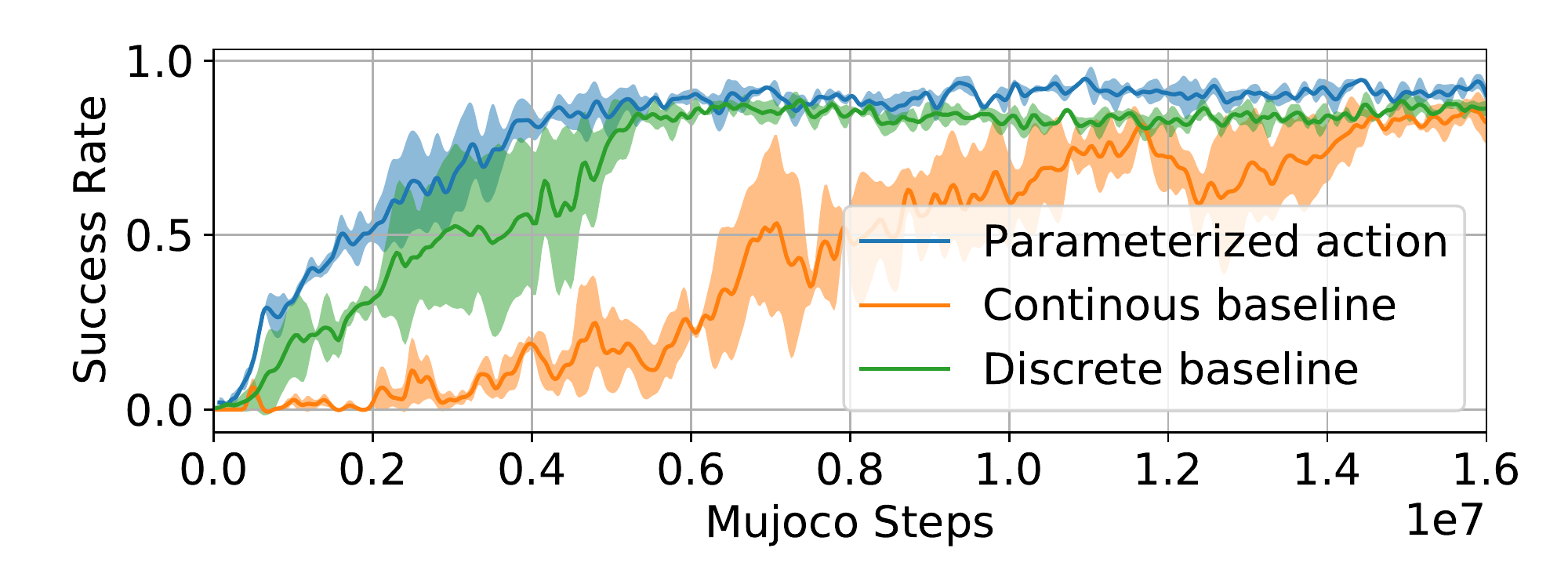}
    \caption{Learning curve comparison}
    \label{fig:learning_curve}
\end{figure}

Fig~\ref{fig:learning_curve} depicts the learning curves. Compared with the two baselines, our proposed approach uses fewer rollout steps to converge, suggesting that learning primitives with continuous parameters can simplify the task and is more efficient. Furthermore, although the training efficiency is similar to the discrete primitive baseline, using parameterized primitive achieves a higher success rate, which agrees with our previous discussion. The detailed success rates are shown in Table~\ref{tab:sim_success_rate}. Note that our proposed TS-MP-DQN achieves a higher success rate than the discrete primitive baseline even with a short horizon, which also suggests our method has shorter execution time.

\begin{table}[http]
    \centering
    \begin{tabular}{c| cccc }
               & TS-MP-DQN & MPDQN & Discrete& Continuous\\
               \cline{1-5}
         Square    &$\bm{96.7\%}$&$93.3\%$&$60.0\%$&$26.7\%$\\
         Triangle  &$\bm{93.3\%}$&$83.3\%$&$46.7\%$&$36.7\%$\\
         Pentagon  &$\bm{93.3\%}$&$87.7\%$&$53.3\%$&$6.7\%$\\
         Triangle(Trans) &$\bm{96.7\%}$&$83.3\%$\\
         Pentagon(Trans) &$\bm{100\%}$&$93.3\%$\\
         \cline{1-5}
         Round Peg &$\bm{86.7\%}$&$80.0\%$&$40.0\%$&$20.0\%$\\
         Waterproof&$\bm{100\%}$&$73.3\%$&$6.7\%$&$0\%$\\
         Ethernet  &$\bm{86.7\%}$&$80.0\%$&$43.3\%$&$10.0\%$\\
    \end{tabular}
    \caption{Success rate evaluation in the real-world. Each method is evaluated for 30 trials on one task. (Trans) indicates that primitives are obtained by the transfer learning. }
    \label{tab:real_success_rate}
\end{table}

\subsubsection{Transfer to Different Tasks}

\hfill

Since primitives are abstractions of element actions and encodes human prior knowledge, we hypothesize that they are more general representations of robot skills and can be transferable between different tasks. Thus, we transfer the learned primitives on the square peg-hole to two other assembly tasks: triangle and pentagon peg-holes.

We first directly transfer policies learned from square peg-hole to other shapes. From Table~\ref{tab:sim_success_rate}, we observe TS-MP-DQN outperforms all three baselines, which shows that the learned primitives are more robust and generalizable for different peg shapes. We then evaluated the fine-tuning performance. As a result, although the actor network is learned from scratch, fine-tuning with online transition data improves the success rate for the triangle peg-hole and achieves similar performance with direct transfer on the pentagon peg-hole.

\subsection{Real Robot Experiments}
\subsubsection{Real Robot Setup}

\hfill

A FANUC LR Mate 200iD robot is used in the real-world experiment. The external force is measured by the F/T sensor and the robot is controlled by an impedance controller.

Fig~\ref{fig:assembly_tasks}(b), (c) shows two sets of real-world peg-hole tasks. We use the same peg-hole shapes as the simulation for the first set to test the policy performance of the RL on the square peg and the transfer learning for two other shapes. The second set includes more realistic robot assembly tasks such as tight round peg-hole, waterproof connector, and the Ethernet connector to test whether our method can generalize to these unseen tasks. Real world experiments have same uncertainty and initial pose settings with simulation. All the insertion primitives are learned in the simulation and directly applied to the real robot without fine-tuning. The details of the peg-hole tasks can be found in Table~\ref{tab:peghole_detail}. 

\begin{table}[http]
    \centering
    \begin{tabular}{c|c|c|c}
         Peg-hole& Hole size & Clearance & Manufacturing \\
         \cline{1-4}
         Square& 51.42 mm& 1.40 mm& 3D Printing\\
         Triangle& 54.50 mm& 1.62 mm& 3D Printing\\
         Pentagon& 57.81 mm& 1.29 mm& 3D Printing\\
         Round Peg& 25.36 mm& 0.05 mm& Machined\\
         Waterproof Connector& & & Purchased\\
         Ethernet Connector& & & Purchased\\
         \cline{1-4}
    \end{tabular}
    \caption{Details of assembly tasks on the real robot}
    \label{tab:peghole_detail}
\end{table}

\subsubsection{Sim-to-real transfer to real robots}

\hfill

First, we focus on the first set of peg-in-hole tasks, which have the same peg-hole shapes as the simulation. As shown in Table~\ref{tab:real_success_rate}, the insertion primitives learned by our proposed method achieve high success rates (over $93.3\%$) by directly sim-to-real transfer on three peg-hole tasks and outperforms all baselines. These values are sightly higher than the simulation and we hypothesize that the extra compliance offered by the impedance control helps the insertion process. Comparing with the continuous control baseline, two primitive-based methods (ours and discrete primitives) are more robust to sim-to-real transfer because using primitives as action improves the generalizability, which is also suggested in \cite{vuong2020learning}. Moreover, benefiting from learning continuous primitive parameters, the primitives obtained by our proposed method are more efficient and have significantly higher success rates than using discrete primitives. One thing to note is that, in our real-world experiment, the performance of discrete primitives baseline is lower than the original paper reported. The reason is that we have higher uncertainty on the hole pose, and the trend of decreasing success rate can also be noticed in \cite{vuong2020learning} when increasing the uncertainty level.

We also tested insertion primitives learned by the transfer learning in simulation. We noticed that transfer learning on new assembly tasks improves the overall success rate than direct transfer. Furthermore, our proposed TS-MP-DQN algorithm outperforms pure MP-DQN in both triangle and pentagon tasks.
\subsubsection{Generalizing to Unseen Peg-hole Shapes}

\hfill

We then evaluate our method on the second set of assembly tasks: tight round peg-hole, waterproof connector, and Ethernet connector. These tasks are challenging because they involve complex peg-hole shapes, and the clearance is much smaller. Thus, we made two modifications: First, the insertion primitives are now learned on a tighter square peg-hole in simulation. Second, we increase the impedance gains for insertion primitive to apply a larger force during the insertion. 

As depicted in Fig~\ref{fig:Snapshots}(b), the learned policy also follows the ``tilt and translate" way to find the hole location, then adjusts the peg orientation to align, and finally uses an insert primitive to finish the task. Although the insertion primitives are learned in the simulation with simple square-shaped peg and hole, they can be directly transferred to realistic, challenging robotic assembly tasks with promising success rate, as shown in Table~\ref{tab:real_success_rate}. This result indicates that our approach has outstanding generalizability and is robust to different task settings.
 


\section{CONCLUSIONS}\label{conclusion}
In this paper, we innovatively formulated robot insertion primitives as parameterized actions, hybrid actions consisting of discrete primitive types and continuous primitive parameters, and reinforcement learning is utilized to learn such insertion primitives. Since primitive is an abstraction of element robot commands, our intuition is that it can reduce the task complexity and is more robust to transfer. Furthermore, comparing with discrete primitives baseline, which can only choose primitive parameters from a discrete set, the primitive parameters in our method are learned from a continuous space and have an infinite number of possible choices. Therefore, the primitives in our proposed method is more flexible and has better performance. To learn primitives in the parameterized action space, we also proposed Twin-Smoothed Multi-pass Deep Q-Network (TS-MP-DQN), an advanced version of MP-DQN with twin Q-network and policy smooth to reduce the Q-value overestimation. 

For validation, we conducted experiments in simulation and real-world and found the policy obtained by our method in simulation can transfer to real-robot tasks with high success rates while outperforming all baselines. Moreover, the proposed method can generalize to challenging and unseen tasks such as tight round peg-hole and electric connectors.


For the future work, we would like to include vision information into the insertion process to offer the initial pose estimation. Furthermore, we also want to extend our method to other tasks such as bimanual and deformable object manipulation. 
\addtolength{\textheight}{-12cm}   






\bibliographystyle{IEEEtran}
\bibliography{IEEEabrv1}
\end{document}